\PassOptionsToPackage{table}{xcolor}
\documentclass{article} 
\usepackage{tencent_ailab_tech_report}
\usepackage[colorlinks = true,
            linkcolor = blue,
            urlcolor  = blue,
            citecolor = blue,
            anchorcolor = blue]{hyperref}   

\usepackage{microtype}
\usepackage{xurl}
\usepackage{booktabs}
\usepackage{enumitem}
\usepackage{multicol}
\usepackage{CJKutf8}
\usepackage{amsmath}
\usepackage{siunitx}
\usepackage{floatflt}
\usepackage{graphicx}
\usepackage{booktabs}
\usepackage{wrapfig}
\usepackage{authblk}
\usepackage{lipsum}
\usepackage{algorithm}
\usepackage{algorithmicx}
\usepackage{algpseudocode}
\usepackage{microtype}
\usepackage{subfigure}
\usepackage{multirow}
\usepackage{booktabs} 
\usepackage{pifont}  
\usepackage{amssymb} 
\usepackage{makecell}
\usepackage[colorinlistoftodos]{todonotes}

\usepackage{array}
\usepackage{amsmath}
\usepackage{amssymb}
\usepackage{mathtools}
\usepackage{amsthm}

\usepackage[capitalize,noabbrev]{cleveref}
\usepackage{adjustbox} 

\usepackage[utf8]{inputenc}
\usepackage[T1]{fontenc}
\usepackage{caption} 
\usepackage{adjustbox} 
\usepackage{arydshln}
\usepackage{fontawesome5}

\usepackage{tcolorbox}
\tcbuselibrary{skins,breakable}

\tcbuselibrary{skins}

\usepackage{soul} 

\definecolor{tred}{RGB}{251, 130, 132}
\definecolor{torange}{RGB}{247, 162, 116}
\definecolor{tyellow}{RGB}{251, 218, 140}
\definecolor{tgreen}{RGB}{127, 204, 181}
\definecolor{tblue}{RGB}{89, 177, 215}
\definecolor{insightblue}{RGB}{162, 210, 255}
\definecolor{questionred}{RGB}{255, 175, 204}

\newcommand{\yes}{\ding{51}}
\newcommand{\no}{\ding{55}}

\newcommand{\tabincell}[2]{\begin{tabular}{@{}#1@{}}#2\end{tabular}} 

\colmfinalcopy

\title{Cognitive Kernel-Pro: A Framework for Deep Research Agents and Agent Foundation Models Training}

\author{%
Tianqing Fang\thanks{Equal Core Contributors}\ , Zhisong Zhang$^*$, Xiaoyang Wang, Rui Wang, Can Qin, Yuxuan Wan, Jun-Yu Ma, Ce Zhang, Jiaqi Chen, Xiyun Li, Yonglin Wang, Jingchen Ni, Tianshi Zheng, Chun Chen, \quad Wenhao Yu, Zhenwen Liang, Hongming Zhang, Haitao Mi, Dong Yu

\vspace{3pt}
Tencent AI Lab\\
\vspace{0.5pt}
\hspace{-10pt}\faGithub ~\url{https://github.com/Tencent/CognitiveKernel-Pro}  \\
\vspace{1pt}
}

\begin{document}

\maketitle

\begin{figure}[h]
    \vspace{-3em}
    \centering
    \includegraphics[width=\textwidth]{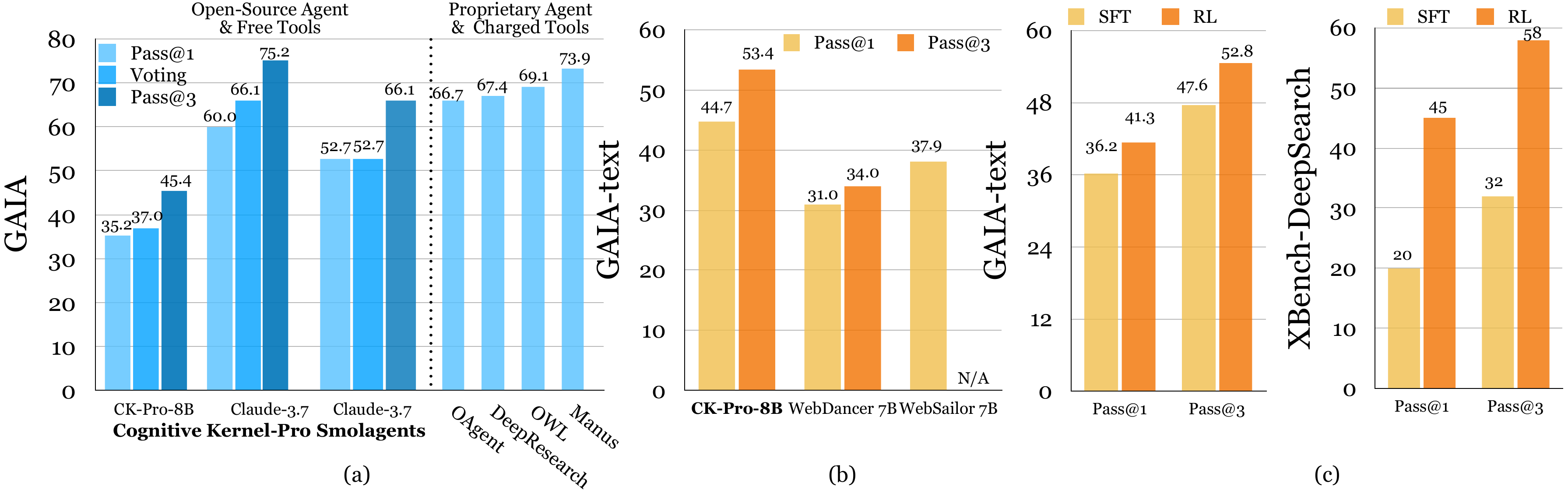} 
    \vspace{-1em}
    \caption{(a) Performance comparison on the full GAIA development set ($n$=165). The left panel presents results from our Cognitive Kernel-Pro framework, utilizing our Qwen3-8B SFT model and Claude-3.7-sonnet as foundation models with exclusively free tools. The right panel displays Pass@1 scores for proprietary agents and systems employing paid tools. (b) Performance on the text-only GAIA subset ($n$=103), demonstrating our 8B model's superiority over 7B models in the WebDancer/WebSailor family ($\sim$2\% higher Pass@1, over 10\% higher Pass@3). (3) Performance after reinforcement learning, over SFT baselines on GAIA-text and XBench-DeepSearch. }
    \label{fig:overview}
\end{figure}

\begin{abstract}

General AI Agents are increasingly recognized as foundational frameworks for the next generation of artificial intelligence, enabling complex reasoning, web interaction, coding, and autonomous research capabilities. 
However, current agent systems are either closed-source or heavily reliant on a variety of paid APIs and proprietary tools, limiting accessibility and reproducibility for the research community. 
In this work, we present \textbf{Cognitive Kernel-Pro}, a fully open-source and (to the maximum extent) free multi-module agent framework designed to democratize the development and evaluation of advanced AI agents. 
Within Cognitive Kernel-Pro, we systematically investigate the curation of high-quality training data for Agent Foundation Models, focusing on the construction of queries, trajectories, and verifiable answers across four key domains: web, file, code, and general reasoning. 
Furthermore, we explore novel strategies for agent test-time reflection and voting to enhance agent robustness and performance. 
We evaluate Cognitive Kernel-Pro on GAIA, achieving state-of-the-art results among open-source and free agents. 
Notably, our 8B-parameter open-source model surpasses previous leading systems such as WebDancer and WebSailor, establishing a new performance standard for accessible, high-capability AI agents.

\vspace{0.4em}

\footnotesize{
\textbf{Note:} The term Cognitive Kernel~\citep{cognitive_kernel} refers to the core computational framework of the agent, designed to emulate the cognitive processes of the human mind.
}
\end{abstract}

\newpage

\begin{figure}[h]
    \centering
    \includegraphics[width=0.8\textwidth]{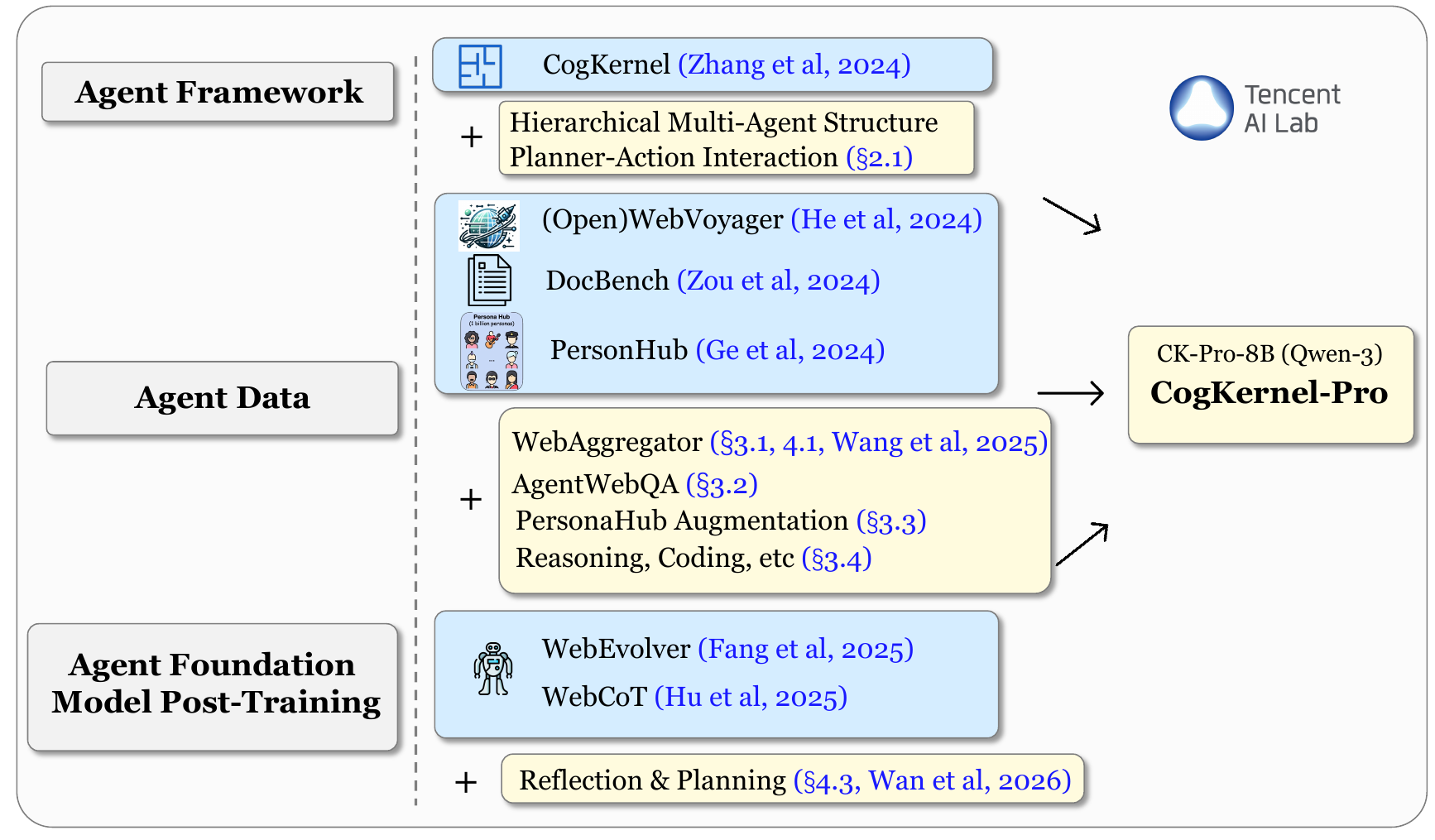} 
    \caption{ Technical roadmap showcasing prior innovations from Tencent AI Lab (Cognitive Kernel;~\citealp{cognitive_kernel}, WebVoyager;~\citealp{webvoyager}, etc) and their integration to \textbf{Cognitive Kernel-Pro} via three core components, agent framework development, agent data construction, and agent foundation model training. Yellow blocks highlight novel contributions in this work and the corresponding section numbers.}
    \label{fig:roadmap}
\end{figure}

\section{Introduction}

The rapid advancement of Deep Research Agents~\citep{manus, deep_research} has transformed the landscape of automated knowledge discovery and problem-solving. 
These agents, powered by large language models (LLMs) and vision-language models (VLMs), excel in tasks such as coding, web navigation, file processing, and complex reasoning. 
However, efforts toward fully open-source agent frameworks~\citep{smolagents, webdancer, websailor} remain limited. 
Existing open-source implementations~\citep{oagent, owl} often rely on proprietary tools like Jina Reader, FireCrawl, or Chunkr to achieve competitive performance, creating barriers to accessibility and reproducibility, or lack of multimodal or general agentic abilities~\citep{webdancer, websailor}.
This dependency on paid tools underscores the need for a robust, \textbf{\textit{fully}} open-source framework that maximizes the inherent capabilities of LLMs and VLMs without external dependencies.

To address this gap, we propose \textbf{Cognitive Kernel-Pro}, a multi-module, hierarchical agent framework designed to facilitate fully open-source agent development. 
Cognitive Kernel-Pro leverages Python code as its action space, harnessing the full reasoning and code-generation potential of modern LLMs. 
The framework adopts a modular architecture, featuring a main agent that orchestrates specialized sub-agents for web navigation, file handling, and tool invocation. 
Each module operates independently, ensuring modularity and extensibility while simplifying the collection of task-specific training data. 
By minimizing reliance on proprietary tools, Cognitive Kernel-Pro emphasizes the intrinsic capabilities of Agent Foundation Models.

In addition to the framework, we introduce a comprehensive training recipe tailored for Cognitive Kernel-Pro, covering diverse domains such as web navigation, file processing, code generation, and reasoning. Our approach includes the construction of verifiable agent query-answer pairs, ensuring high-quality training data. To enhance data collection, we incorporate intermediate process hints and employ hint-based rejection sampling, which significantly improves the quality and relevance of the collected data. This structured training methodology enables Cognitive Kernel-Pro to achieve robust performance across diverse tasks while maintaining full open-source compatibility.

Furthermore, we explore inference-time optimization techniques to address the inherent randomness in tasks like web browsing. To mitigate variability, we propose a pipeline that integrates retry mechanisms and ensemble-based multi-run strategies. This approach enhances the reliability and consistency of Cognitive Kernel-Pro’s performance, particularly in dynamic and unpredictable environments. By combining a modular framework, a robust training recipe, and optimized inference strategies, Cognitive Kernel-Pro sets a new standard for open-source agent development, paving the way for accessible and reproducible advancements in agent-based research.



\section{Cognitive Kernel-Pro Framework}\label{sec:ck_arch}

\begin{figure}[t]
    \centering
    \includegraphics[width=\textwidth]{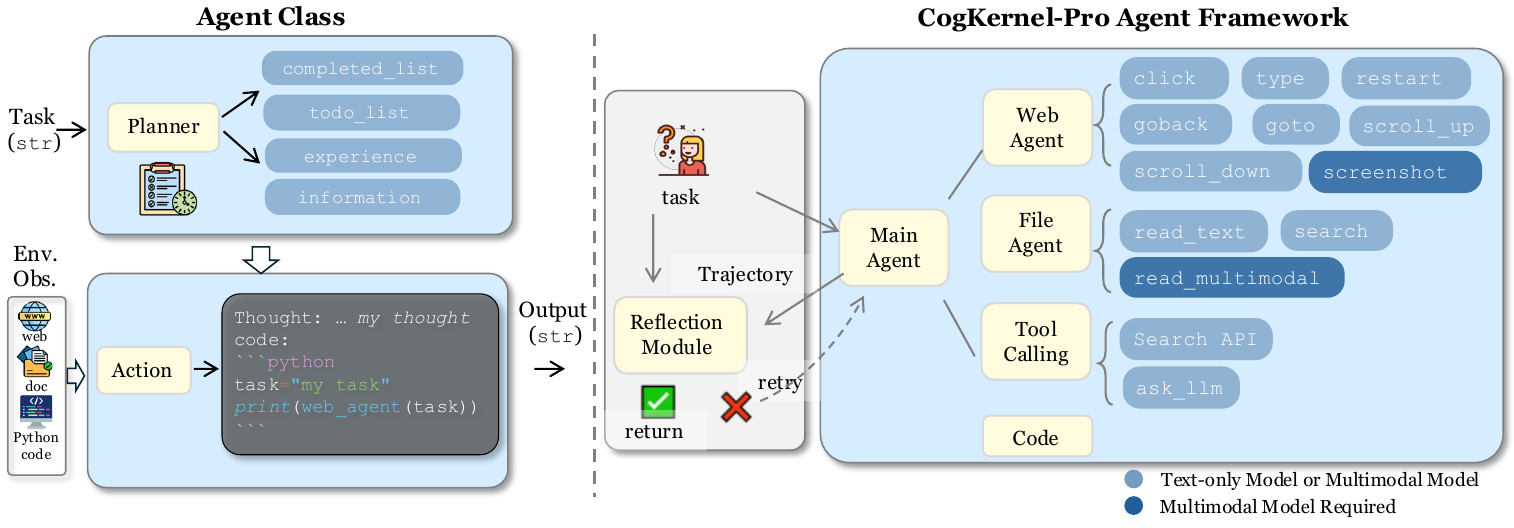} 
    \caption{Overview of the Cognitive Kernel-Pro Agent Framework. The left panel illustrates the functionality of agent class, where the main agent, web agent, and file agent inherit from the common base class. 
    The planner maintains a state dictionary containing `completed\_list', `todo\_list', `experience', and `information' (\S \ref{sec:ck_arch}). 
    The action generator produces Python code as a code agent or invokes predefined functions of sub-agents, such as the web agent, as well as other built-in tools. 
    The right panel illustrates the hierarchical structure of Cognitive Kernel-Pro, listing all functions defined by each agent. 
    Additionally, a standalone reflection module is included to assess task completion; if the task is incomplete, the agent will retry (\S \ref{sec:reflection}). 
    The agent foundation model behind each module/sub-agent is the same.
    }
    \label{fig:agent_framework}
\end{figure}

We present an overview of the Cognitive Kernel-Pro framework in Figure~\ref{fig:agent_framework}.


We adopt a two-tier multi-module framework in our agent implementation. This framework consists of a main agent, responsible for task decomposition, sub-task delegation, and information aggregation, tool calling, code generation, as well as several sub-agents, whose objective is to solve the sub-tasks assigned by the main agent. 
Both the main agent and sub-agents inherit from the same base class, where the input is a \textit{task string}, the output is a \textit{response string}, and intermediate actions are executed as \textit{Python code}.

\paragraph{Main-Agent.} The main agent directly manages the problem-solving process towards achieving the overall goal. It decomposes the original complex tasks into manageable sub-tasks and assigns these to sub-agents as needed. Upon receiving responses from the sub-agents, the main agent aggregates the information and continues with the main procedure. Notably, the main agent does not possess specialized skills such as web browsing or file processing; only the sub-agents (e.g., the web agent and the file agent) are equipped with such capabilities. Nevertheless, the main agent is aware of the functionalities of the sub-agents and is in charge of delegating appropriate sub-tasks accordingly.

\paragraph{Sub-Agents.} The sub-agents are equipped with specialized skills that are essential for a general-purpose task-solving agent system. Each sub-agent follows a multi-step procedure similar to that of the main agent but is enhanced with specialized actions that enable direct interaction with specific resources. In our system, we primarily include the following two sub-agents:
\begin{itemize}
    \item \textbf{Web Agent}. The web agent is equipped with a browser and can navigate live web pages to collect relevant and time-sensitive information. We implement an autonomous web browser using \texttt{playwright}, which provides both the accessibility tree and the screenshot of the current web page. The web agent makes decisions based on the current web page's observations. We adopt typical web agent actions, including ``click'', ``type'', ``scroll'', ``wait'', ``goback'', ``restart'', ``goto'', ``save'', ``stop'', ``screenshot''. Here, ``save'' refers to explicitly saving a web file to a local path for the file agent to process, while ``stop'' denotes terminating the navigation process due to task completion or unrecoverable errors. 
    ``screenshot'' is a special function to turn on screenshot mode to call a multimodal language model to process the image. If this function is not called, the default input to the agent foundation model is the text-only accessibility tree of the current webpage.
    \item \textbf{File Agent}. The file agent is designed to process a variety of file types, such as PDF files (.pdf), spreadsheet files (.xlsx, .xls, .csv), and image files ('.png', '.jpg', '.gif', etc.). Inspired by the web agent, we adopt a similar task-solving process. To manage potentially large files, we split each file into pages (using specialized tools for each file type) and allow the file agent to read a subset of pages at a time. Correspondingly, the action space includes ``load\_file'', ``read\_text'', ``read\_screenshot'', ``search'' and ``stop''. 
    Here, the file agent can decide whether to read the screenshot of certain pages or only the text, with the screenshot mode being essential for image-based tasks.
\end{itemize}

While we do not have a standalone \textbf{Code Agent} in the system, 
every sub-agent is a code agent since the output action of every agent is essentially python code. 
For example, every agent can generate Python code to perform calculation or other reasoning tasks that can be solved by code generation and execution.

In addition to these sub-agents, our framework is flexible and can be extended to support more sub-agents with specialized skills. The design of this two-tier multi-module framework enables the decoupling of the main agent's task-solving procedure from the detailed sub-task execution of the sub-agents, providing a flexible and adaptable system capable of supporting a wide range of scenarios.

\begin{table}[t]
\vspace{-2em}
\centering
\small
\setlength{\tabcolsep}{1.3mm}{
\scalebox{1.0}{
\begin{tabular}{l|c c c |c c c}
\toprule
\midrule
\multirow{2}{*}{Agent} &  \multirow{2}{*}{\tabincell{c}{Open\\ Framework}} & \multirow{2}{*}{\tabincell{c}{Open-source\\ Model}} & \multirow{2}{*}{\tabincell{c}{No Proprietary\\Tool (excl. Google)}}  & \multicolumn{3}{c}{Agent Ability} \\   
 & & &  & \quad Web \quad & \quad File \quad & \quad Code \quad \\
\midrule
Deep Research & \no & \no & \no & \yes & \yes & \yes \\
OWL & \yes & \no & \no & \yes & \yes & \yes \\
OAgents & \yes & \no & \no & \yes & \yes & \yes \\
WebDancer & \yes & \yes & \yes & \yes & \no & \no \\
WebSailor & \yes & \yes & \yes & \yes & \no & \no \\
\midrule
\textbf{Cognitive Kernel-Pro} & \yes & \yes & \yes & \yes & \yes & \yes \\
\midrule
\bottomrule
\end{tabular}
}
\caption{Feature Comparison of AI Agent Frameworks. Google Search API (which, can be easily switched to free APIs such as DuckDuckGo if needed) is excluded when comparing proprietary tools because it's a must in search-related tasks. Note: WebDancer and WebSailor support PDF fetching but lack general file agent capabilities}
\label{tab:paid_tool}}
\end{table}

\paragraph{Tool Calling}

Our system utilizes minimum paid tools. We use Google Search API to return search results, wrapped as a function ``simple\_web\_search''. Besides, we only use ``ask\_llm'' as an additional function to directly let the base agent foundation model to answer a question.
Despite Google Search API is not free but it is required by most information seeking agents. Other than that, we do not use any proprietary tools, as illustrated in Table~\ref{tab:paid_tool}.

The detailed implementations are presented in the Appendix~\ref{appx:agent_implementation}.

\begin{table}[t]
\centering
\small
\begin{tabular}{l| >{\raggedright\arraybackslash}p{5.0cm} >{\raggedright\arraybackslash}p{4cm} r r}
\toprule
\midrule
\rowcolor{cyan!20} \textbf{Type} & \textbf{Data Name} & \textbf{Data Type} & \textbf{\#Query} & \textbf{\#Steps} \\
\midrule
\cellcolor[RGB]{255, 255, 140}{Web} & \cellcolor[RGB]{255, 255, 150}{OpenWebVoyager~\citep{openwebvoyager}} & \cellcolor[RGB]{255, 255, 180}{Web Browsering} & \cellcolor[RGB]{255, 255, 200}{1,259} & \cellcolor[RGB]{255, 255, 220}{9,098} \\
 \cellcolor[RGB]{255, 255, 140}{\quad}& \cellcolor[RGB]{255, 255, 150}{Multihop URLQA (\S \ref{sec:urlqa})} & \cellcolor[RGB]{255, 255, 180}{Web Information Seeking} & \cellcolor[RGB]{255, 255, 200}{4,225} & \cellcolor[RGB]{255, 255, 220}{25,589} \\
 \cellcolor[RGB]{255, 255, 140}{}& \cellcolor[RGB]{255, 255, 150}{AgentWebQA (w/ hint) (\S \ref{sec:agentqa})} & \cellcolor[RGB]{255, 255, 180}{Web Information Seeking} & \cellcolor[RGB]{255, 255, 200}{2,721} & \cellcolor[RGB]{255, 255, 220}{32,231} \\
 \cellcolor[RGB]{255, 255, 140}{\quad}& \cellcolor[RGB]{255, 255, 150}{PersonaHub-Aug (\S \ref{sec:personahub})} & \cellcolor[RGB]{255, 255, 180}{(No Ground Answer)} & \cellcolor[RGB]{255, 255, 200}{1,000} & \cellcolor[RGB]{255, 255, 220}{2,088} \\
 \cellcolor[RGB]{255, 255, 140}{\quad} & \cellcolor[RGB]{255, 255, 150}{WebWalkerQA~\citep{WebWalker}} & \cellcolor[RGB]{255, 255, 180}{Web Information Seeking} & \cellcolor[RGB]{255, 255, 200}{1,904} & \cellcolor[RGB]{255, 255, 220}{18,116} \\
 \midrule
 \cellcolor[RGB]{255, 170, 180}File & \cellcolor[RGB]{255, 182, 193}{DocBench~\citep{zou2024docbench}} & \cellcolor[RGB]{255, 200, 200}{.pdf} & \cellcolor[RGB]{255, 220, 220}{300} & \cellcolor[RGB]{255, 235, 235}{1,566} \\
  \cellcolor[RGB]{255, 170, 180}& \cellcolor[RGB]{255, 182, 193}{TableBench~\citep{wu2025tablebench}} & \cellcolor[RGB]{255, 200, 200}{.csv, .xlsx} & \cellcolor[RGB]{255, 220, 220}{1,000} & \cellcolor[RGB]{255, 235, 235}{9,482} \\
 \midrule
  \cellcolor[RGB]{210, 215, 245}Reasoning & \cellcolor[RGB]{215, 215, 250}{NumiaMath~\citep{numina_math_7b}} & \cellcolor[RGB]{220, 220, 250}Math Reasoning & \cellcolor[RGB]{240, 240, 250}616 & \cellcolor[RGB]{244, 244, 250}524 \\
 \cellcolor[RGB]{210, 215, 245}  & \cellcolor[RGB]{215, 215, 250}BAAI/TACO~\citep{li2023taco} & \cellcolor[RGB]{220, 220, 250}Code/Puzzle & \cellcolor[RGB]{240, 240, 250}225 & \cellcolor[RGB]{244, 244, 250}730 \\
 \cellcolor[RGB]{210, 215, 245}& \cellcolor[RGB]{215, 215, 250}RiddleSense~\citep{lin-etal-2021-riddlesense} & \cellcolor[RGB]{220, 220, 250}Riddle/Puzzle & \cellcolor[RGB]{240, 240, 250}179 & \cellcolor[RGB]{244, 244, 250}165 \\
 \cellcolor[RGB]{210, 215, 245}& \cellcolor[RGB]{215, 215, 250}LogiCoT~\citep{liu-etal-2023-logicot} & \cellcolor[RGB]{220, 220, 250}Logical Reasoning & \cellcolor[RGB]{240, 240, 250}1,400 & \cellcolor[RGB]{244, 244, 250}1,400 \\
\midrule
\bottomrule
\end{tabular}
\caption{Summary of the training recipe.}\label{tab:training_data}
\end{table}



\section{Cognitive Kernel-Pro Agent Foundation Model Supervised Fine-Tuning}\label{sec:data}

\subsection{Overall Data Recipe}

The overall training data recipe is presented in Table~\ref{tab:training_data}. We divide the ability of deep research agent into three types, Web, File, and Reasoning. For each of the category, we either convert the existing benchmarks to the format that we need or construct new deep research queries (\S\ref{sec:urlqa}, \S\ref{sec:agentqa}, and \S\ref{sec:personahub}).

\subsection{Multi-hop Web Search Data Construction} \label{sec:urlqa}

The data synthesis procedure here aims to create diverse and complex multi-hop information-seeking QA pairs grounded in web pages. 
We expect the constructed questions requiring information cannot be obtained without a retrieval process. 
To cover multiple domains, we first collect a seed URL set by searching for topic-diverse texts from several datasets using the commercial API of Google. 
Then, an agent traverses and browses web pages starting from these seed URLs with the designed prompt and examples, gathering information and composing questions accordingly.

\begin{figure}
    \centering
    \includegraphics[width=1\linewidth]{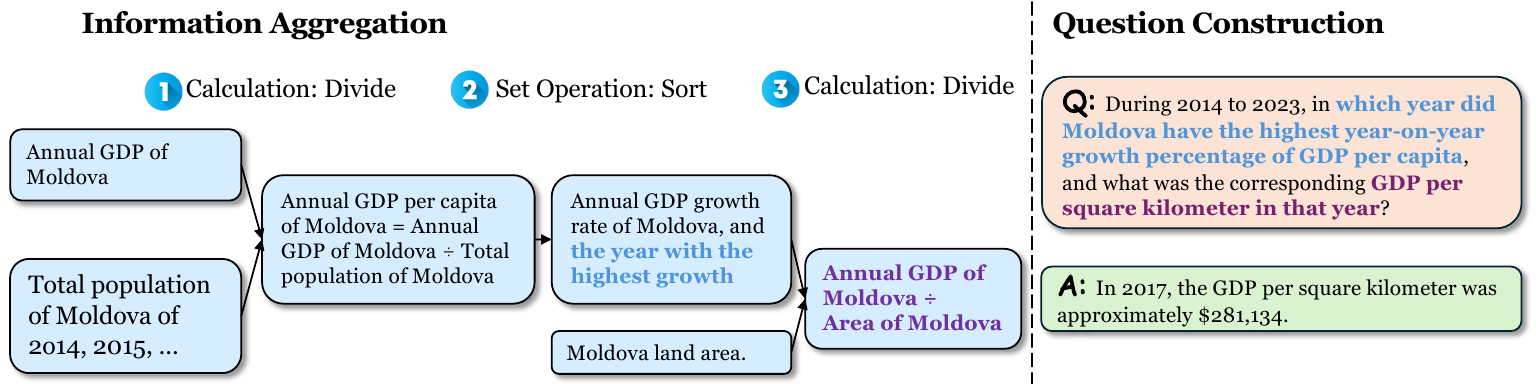}
    \caption{Illustration of information aggregation in the creation of URLQA.}
    \label{fig:info-agg}
\end{figure}

Additionally, to simulate varied task intents, we add the principals and several examples in the prompt, constraining that the answer must be derived through information aggregation operations, as shown in Figure~\ref{fig:info-agg}.
The composition rules are specially designed for different forms of information sources, such as math calculation for numbers, sorting for candidate sets, data analysis for tables.

A comprehensive extension of this pipeline is presented as a separate paper in WebAggregator~\citep{webaggregator}.

\subsection{Agent Exploration-based Data Construction} \label{sec:agentqa}

The data synthesizing process can be viewed as a specialized task, for which we can re-use our existing agent framework. In this context, all sub-agents remain unchanged, while the main agent is adapted specifically for data synthesis. The overall procedures and mechanisms are largely consistent with those of our general-purpose framework. We adopt several modifications to further tailor the framework for the data synthesizing process.

\paragraph{Prompt Adjustments.}
The data synthesizing process operates in a way that is essentially the reverse of the ordinary task-solving procedure. In ordinary task-solving, the agent is provided with a question and aims to search for the answer. In contrast, data synthesizing requires the agent to construct the query itself by integrating pieces of information gathered throughout the exploration process. To accommodate this use case, we retain the core mechanisms of the main agent but revise its prompts to address the unique requirements of data synthesis. In particular, we instruct the agent to construct complex queries by combining information from multiple verifiable sources. 

\paragraph{Topic Sampling.} To allow a diverse set of interesting queries that are synthesized, before the agent-based data synthesizing procedure, we first generate an overall topic for each query to be synthesized. Using a self-instruct based method, we use an LLM to generate broad and interesting topics with verifiable sources of truth. We provide several seed examplse and let the LLMs to generate more. After generation, we adopt a diversity-based sampling procedure to sample a diverse subset of topics for our actual query synthesizing process.

\paragraph{Hint-based Training Trajectory Sampling.}
The query synthesizing procedure yields not only the constructed queries but also all associated intermediate and final results. We observe that, during trajectory sampling for training, providing these intermediate results as hints to the task-solving agent significantly improves the success rate for training data collection. To enhance this, we augment the queries with additional textual hints. It is important to note that this augmentation is employed only during training data collection, where we can assume that answers are available. Once the trajectories are obtained, all such hints are removed from the model inputs and outputs prior to the actual training process. The hints are wrapped between \texttt{<secret>} and \texttt{</secret>} during sampling, and will be removed during training.

The prompt of Data Synthesizing Requirements, Topic Sampling, and Hint-based Training Trajectory Sampling are presented in Appendix~\ref{appx:agent_query_construction_details}.

\subsection{Persona Hub-based Data Augmentation}\label{sec:personahub}


\paragraph{Persona-triggered Query Synthesis.}
PersonaHub~\citep{ge2024scaling} provides an effective strategy to synthesize large-scale diverse queries for various LLM tasks like math, logic reasoning, instruction following, etc. In this work, we explore to utilize Persona Hub to synthesize training queries for the deep research agent task. Seeded with a manually crafted deep research question and its corresponding persona as the in-context example, we utilize an LLM to generate a synthetic deep research query given a synthetic persona from Persona Hub. Though with limited number of manually crafted deep research questions, we can easily scale up the synthesis of diverse deep research queries by using more personas from Persona Hub as triggers.

\paragraph{Trajectory Sampling and Validation.}
The main challenge of Persona Hub-based data augmentation for deep research agent task is the lack of ground truth answer for the synthetic query. To tackle this challenge, we conduct cross-validation of the trajectory outcomes from different agent systems and include 1k synthetic queries with their trajectories from Cognitive Kernel-Pro agent system into our training set, as detailed in Table~\ref{tab:training_data}. Our ablation study suggests a small number of training trajectories from Persona Hub-based data augmentation can effectively improve performance. On the other hand, manual validation and response annotation of these synthetic queries are considerable but not yet included in this work.

\subsection{Reasoning Data Construction}

We refine several existing reasoning datasets relevant to general agents, including NumiaMath (applied mathematical reasoning), LogiCoT (logical reasoning), TACO (code reasoning), and RiddleSense (puzzles and riddles). For TACO, we extract input/output pairs from the task descriptions and construct code agent queries by concatenating the task description with the input case, using the corresponding output as the expected gold answer. For the other datasets, we directly transform the question-answer pairs into the input-output format compatible with the ``ask\_llm'' function.

\subsection{Trajectory Sampling}

For all constructed query-answer pairs, we utilize \texttt{gpt-4.1} as the foundational backbone model within our Cognitive Kernel-Pro framework to generate agent trajectories. Subsequently, we apply rejection sampling using similarity-based matching, facilitated by the `cot\_qa` of LangChain, with \texttt{gpt-4.1} as the backbone model. 
For hint-based sampling, we exclude all hints enclosed within \texttt{<secret>} and \texttt{</secret>} tags to prevent information leakage. Each query is sampled up to three times until successful completion.

\begin{table}[t]
\centering
\small
\begin{tabular}{lp{2.4cm}p{3.0cm}cccc}
\toprule
\midrule
\textbf{Framework} & \textbf{Agent Model} & \textbf{Paid Tools} & \textbf{Avg.} & \textbf{Level 1} & \textbf{Level 2} & \textbf{Level 3} \\
\midrule
\multicolumn{7}{l}{\cellcolor[RGB]{255, 255, 191}{\emph{Closed-source Agent Frameworks}}} \\
\midrule
TraseAgent    
& Claude  & Unknown & 70.30 & 83.02 & 69.77 & 46.15 \\
Deep Research
& Unknown & Unknown & 67.36 & 74.29 & 69.06 & 47.60 \\
h2oGPTe   
& Claude-3.5  & Unknown & 63.64 & 67.92 & 67.44 & 42.31 \\
Desearch         
& GPT-4o & Unknown & 56.97 &  71.70 & 58.14 & 23.08 \\
Manus & Claude, etc. & Unknown & 73.3 & 86.5 & 70.1 & 57.7 \\
\midrule
\multicolumn{7}{l}{\cellcolor[RGB]{140, 210 255}{\emph{Open‐source Agent Frameworks}} }\\
\midrule
\textbf{w/ Paid Tools} \\
\midrule
OWL--Roleplaying & 4o \& o3-mini   &  \multirow{2}{*}{\makecell[l]{Chunkr, FireCrawl,\\ Whisper, o3-mini}} &  58.18 & 81.14 & 54.65 & 23.08 \\
OWL--Workforce  & Claude-3-7  &  & 69.09 & 84.91  & 67.44 & 42.31 \\
OWL--Workforce* & Claude-3-7 & w/o whisper & 60.61 & 73.58 & 62.79 & 26.92 \\
OAgent & \multirow{2}{*}{Claude-3-7} &\multirow{2}{*}{\makecell[l]{Jina Reader, Whisper,\\ Baidu \& Bing API}} & 66.67 & 77.36 & 66.28 & 46.15  \\
\quad -Pass@3 & &  & {73.93} & {83.02} & {74.42} & {53.85} \\
\midrule
\textbf{w/o Paid Tools} \\
\midrule
TapeAgents   & Claude-3-7 & ---  & 55.76 & 71.70 & 53.49 & \underline{30.77} \\
AutoAgent  & Claude-3-5 & --- & 55.15 & 71.70 & 53.40 & 26.92 \\
Magnetic-1 & OpenAI o1 & --- & 46.06 & 56.60 & 46.51 & 23.08 \\
Smolagents & Openai o1 & --- & 49.70 & 54.72 & 53.49 & 26.92 \\ 
Smolagents* & Claude-3-7  &  --- &52.10 &  60.38 & \underline{54.65} & 26.92 \\ 
\quad  - Voting & & --- & 53.99 & 66.04& 52.33 & 33.33\\
\quad  - Pass@3 & & --- & 63.64&75.47 & 61.63 &46.15   \\
\midrule
\textbf{Cognitive Kernel-Pro} &   &  --- & \underline{57.58} & \underline{77.36} & \underline{54.65} & 26.92  \\
-Voting & Claude-3-7 & ---  & 63.64 & 69.81 & 65.12 & 46.15 \\
-Pass@3 &  &  --- &   \textbf{70.91} & \textbf{83.02} & \textbf{68.60} & \textbf{53.85} \\
\textbf{Cognitive Kernel-Pro} & & --- & 32.73 & 43.40 & 32.56 & 11.54   \\
-Voting  & CK-Pro-8B  & --- & 34.54 & 47.17 & 33.72 & 11.54 \\
-Pass@3   & & --- & 38.18 & 50.94 &	38.37 &	11.54	 \\
\midrule
\bottomrule
\end{tabular}
\caption{Performance of various agent frameworks on GAIA dev set ($n$=165). * after agent names indicate our reproduced results. We \textbf{boldface} the best pass@3 performance and \underline{underline} the best pass@1 performance of open-source agent frameworks without paid tools (except for Google Search).}
\label{tab:main_gaia_results}
\end{table}


\subsection{Experiment Setup}

\paragraph{Baselines}
Based on the open-source code of OWL, we reproduced OWL's performance using Claude-3.7-Sonnet in our own environment. All experimental settings followed the default configurations provided by OWL, including the use of corresponding LLM APIs for each agent and the integration of certain paid tools, such as Chunkr and FireCrawl. All agents adopted greedy decoding during their inference, and the maximum number of replanning tries was set to the default value of 2. It should be noted that we did not use the Whisper API, and our network environment was different from that of the original experiments. These factors may have contributed to the reproduced performance being lower than the original results reported by OWL.
As for the implementation of the SmolAgents, our experiment utilizes most of the tools provided by the Open Deep-Research version of SmolAgents and follows its configuration, except that we enhance the web browsing tool with DOM tree parsing to display web structure, enable element clicking, and text input.

\paragraph{Cognitive Kernel-Pro} We only use one paid tool, Google Search API, which is a must for almost all agent frameworks. Claude-3.7 is used as the backbone for supporting the agent framework. We also use our fine-tuned CK-Pro-8B (based on Qwen-3-8B) as the agent foundation model.

\paragraph{Datasets} We use the GAIA dataset~\citep{GAIA} as the evaluation benchmark, a comprehensive suite designed to assess the general intelligence and multi-step reasoning capabilities of AI agents across diverse tasks, including web navigation, question answering, file manipulation, and multimodal processing, making it ideal for evaluating the performance of our Cognitive Kernel-Pro framework.

\begin{table}[t]
\centering
\small
\begin{tabular}{lc|c|cccc}
\toprule
\midrule
\textbf{Framework} & \textbf{Model Size} & \textbf{Avg.} & \textbf{Level 1} & \textbf{Level 2} & \textbf{Level 3} \\
\midrule
WebThinker-Base & 32B & 44.7 & 53.8 & 44.2 & 16.7 \\
WebThinker-RL & 32B & 48.5 & 56.4 & 50.0 & 16.7  \\
Search-o1 & 32B & 28.2 & 33.3 & 25.0 & 0.0  \\
WebDancer & 32B & 40.7 & 46.1 & 44.2 & 8.3  \\
WebDancer & 32B & 51.5 & 61.5 & 50.0 & 25.0   \\
WebSailor & 32B & 53.2 & - & - & -  \\
WebSailor & 72B & 55.4 & - & - & -  \\
WebShaper & 32B & 53.3 & 69.2 & 50.0 & 16.6 \\
WebShaper & 72B & 60.1 & 69.2 & 63.4 & 16.6  \\
\midrule
\midrule
Search-o1 & 7B & 17.5 & 23.1 & 17.3 &  0.0  \\
R1-Searcher & 7B & 20.4 & 28.2 &  19.2 & 8.3  \\
WebDancer & 7B & 31.0 & 41.0 & 30.7 & 0.0  \\
\quad -Pass@ 3 & 7B & 34.0 & - & - & - \\
WebSailor & 7B & 37.9 & - & - & - \\
\midrule
\textbf{Cognitive Kernel-Pro} & 8B &  \underline{43.7} & \underline{56.4} & \underline{42.3} & \underline{8.33}  \\
-Voting  & 8B  & 41.1  & 53.8  & 34.6 & 16.7 \\
-Pass@3   &  8B & \textbf{49.3} & \textbf{61.5} & \textbf{44.2} &	\textbf{16.7} \\
\midrule
\bottomrule
\end{tabular}
\caption{Performance of open-source agent frameworks on the text-only subset of GAIA ($n$=103). We \textbf{boldface} the best Pass@3 performance and \underline{underline} the best pass@1 performance for models with size 7 or 8B. }
\label{tab:gaia_text}
\end{table}

\subsection{Results} 

\paragraph{Full dev set of GAIA} Table \ref{tab:main_gaia_results} shows the performance of various agent frameworks on the complete GAIA dataset, differentiating between closed-source and open-source systems, with the latter grouped by their use of paid tools, and featuring our reproduced results marked with an asterisk (*). Cognitive Kernel-Pro, utilizing Claude-3.7, surpasses Smolagents by 5\% in Pass@1 and 7\% in Pass@3 under identical experimental conditions (e.g., LLM and Search APIs, Internet connectivity), demonstrating its efficacy. Its performance also rivals OWL, which relies on proprietary tools like Chunkr for file processing and FireCrawl for web browsing, underscoring its significant potential.

Additionally, we present results from fine-tuning a Qwen-3-8B model on the trajectories outlined in Section \S\ref{sec:data}, supported by GPT-4.1 for multimodal functions, achieving a Pass@3 score of 38.18\%—a 30\% gap from the state-of-the-art Claude-3.7 model—suggesting considerable scope for future enhancements.

\paragraph{Text-only Subset of GAIA} We present the performance comparisons on the text-only subset of GAIA in Table~\ref{tab:gaia_text}. The major baseline is the 7B version of WebDancer and WebSailor. In addition, we list the performance of 32B and 72B models as a reference in the upper half of the table. We also include the performance of Search-o1~\citep{li2025searcho1}, R1-Searcher~\citep{r1searcher}, and WebThinker~\citep{webthinker} in the table. Cognitive Kernel-Pro under CK-Pro-8B model yield the best pass@1 and pass@3 performance across all levels of GAIA.
\begin{table}[t]
\centering
\small
\begin{tabular}{lp{2.4cm}|c|cccc}
\toprule
\midrule
\textbf{Inference-time Alg.} & \textbf{Inference-time Model}  & \textbf{Avg.} & \textbf{Level 1} & \textbf{Level 2} & \textbf{Level 3} \\
\midrule
w/o Reflection & --- &   27.0 & 35.8 & 27.9 & 7.7  \\
Reflection & CK-Pro-8B &  28.5 & 37.9 & 29.4 & 7.7 \\
Reflection & Qwen-3-32B & 31.5 & 41.5 &	32.5 &	7.7  \\
Reflection & GPT-4.1 & 32.7 & 43.4 & 32.6 & 11.5 \\
\midrule
\bottomrule
\end{tabular}
\caption{Ablations on different backbone LLM used for reflection and voting. }
\label{tab:reflection_ablation}
\end{table}

\begin{table}[t]
\centering
\small
\begin{tabular}{lp{3cm}|c|cccc}
\toprule
\midrule
\textbf{Base Agent Model} & \textbf{MLLM}  & \textbf{Avg.} & \textbf{Level 1} & \textbf{Level 2} & \textbf{Level 3} \\
\midrule
CK-Pro-8B pass@1 & Qwen-2.5-VL-72B & \underline{33.94} & \underline{43.40} & \underline{34.88} & \underline{11.54} \\
CK-Pro-8B pass@1 & GPT-4.1 & 32.67 & \underline{43.40} & 32.56 & \underline{11.54}  \\
\midrule
CK-Pro-8B  pass@3 & Qwen-2.5-VL-72B & 37.56 & 49.06 & \textbf{38.64} & \textbf{11.54} \\
CK-Pro-8B  pass@3 & GPT-4.1 & \textbf{38.12} & \textbf{50.94} &38.37 &	\textbf{11.54}\\
\midrule
\bottomrule
\end{tabular}
\caption{Performance of using CK-Pro-8B as the base agent foundation model and variances of multimodal language model we use. We \underline{underline} the best results on pass@1 and \textbf{boldface} the best performance on pass@3 }
\label{tab:multimodal_ablation}
\end{table}
\paragraph{Ablation Study of Reflection } We present an ablation study of the effect of the reflection module in Table~\ref{tab:reflection_ablation}. Using an open-source model Qwen-3-32B is already good enough, counterparting GPT-4.1. However, if we use our trained CK-Pro-8B model, without being finetuned with reflection ability, there is only marginal improvement. This indicates a future direction of involving the ability of reflection to the training of agent foundation models.

\paragraph{Ablation Study of the Multimodal Language Model}  Table~\ref{tab:multimodal_ablation} presents the impact of using different backbones for our multimodal language model. 
Our results show that replacing Qwen-2.5-VL-72B with GPT-4.1 yields only marginal performance improvements. This suggests that the observed performance gains are not solely because of the use of a more advanced multimodal model like GPT-4.1, as Qwen-2.5-VL-72B achieves comparable results. 
Future work will be developing a fully multimodal language model as the backbone, designed to seamlessly support both text and multimodal inputs.

\section{Cognitive Kernel-Pro Agent Reinforcement Learning}
\subsection{Data Construction and Filtering}
Unlike standard SFT data, RL training necessitates samples characterized by rigorous internal consistency and unambiguous grounding to ensure the derivation of high-fidelity reward signals.
Using the methodology of WebAggregator \citep{webaggregator}, our data samples are structured as <Question, Answer, Reference Solution, Grounded URLs>. 


\paragraph{Multi-stage Quality Filtering}

Starting with the initial dataset, we apply a multi-stage filtering protocol to ensure that every sample is deterministic and verifiable.

We first evaluate whether the provided solution can logically and correctly lead to the reference answer. We utilize an Agent to derive the answer from the reference reasoning steps. 
If the derived prediction from the reference solution is inconsistent with the reference answer, the sample is discarded. This step acts as a rigorous sanity check, filtering out raw noisy data.
Then, a computationally intensive verification where an agent confirms the validity of external sources and the correctness of the reasoning steps.

Moreover, for each candidate sample, we perform four independent rollouts using the CK-Pro SFT checkpoint and retain 400 high-quality samples whose rollouts exhibit partial correctness. Based on this multi-rollout verification, we derive a curriculum signal that captures both verifiability and logical depth. RL training starts with high-consistency samples to provide a denser and more stable reward signal, and then progressively shifts to more complex, lower-consistency tasks.

\paragraph{Structured Subgoal Recovery}

A primary challenge in RL for deep research is the sparsity and delay of rewards caused by exceptionally long reasoning trajectories. 
To facilitate more granular credit assignment, we decompose these long-form solutions into intermediate milestones.
We leverage the rich metadata from the synthesis stage (e.g., URLs, and intermediate states) to reconstruct the reasoning path into a structured subgoal dictionary. Each entry in the dictionary represents a discrete, verifiable milestone with clear success criteria.




\subsection{Subgoal-Conditioned GRPO}

Due to the sparsity and temporal delay of reward signals in long-horizon trajectories, applying reinforcement learning (RL) to deep research agents for complex, multi-step information retrieval is inherently sample-inefficient. To address this challenge, we propose a novel \textbf{subgoal-conditioned GRPO} framework that explicitly separates subgoal-level and task-level supervision, thereby mitigating the ``reward collapse'' phenomenon commonly observed in multi-objective RL. In this section, we formally define the setting and present the proposed optimization method.

\paragraph{Atomic Main-Agent Step Tuple \((h_t)\).}
At each discrete time step \(t\), the interaction of the main agent with the environment is represented by an atomic step tuple
$$h_t = \langle s_t, a_t, \tau_t^{\mathrm{sub}}, k_t, r_t, r^{o} \rangle.$$

Each component is defined as follows:
\begin{enumerate}
    \item \(s_t \in \mathcal{S}\): the main-agent state, comprising the current observation \(O_t\) together with the accumulated memory or context \(m_t\).
    \item \(a_t \in \mathcal{C}\): the main-agent action, consisting of the generated reasoning content, plan, and executable Python code.
    \item \(\tau_t^{\mathrm{sub}}\): the sub-agent trajectory induced by \(a_t\), i.e., the execution trace produced by the lower-level agent after the main agent issues action \(a_t\). Concretely, \(\tau_t^{\mathrm{sub}}\) consists of a sequence of sub-agent state--action pairs \(\langle s^{\mathrm{sub}}_{\ell}, a^{\mathrm{sub}}_{\ell} \rangle\).
    \item \(k_t \in \mathbb{N}\): the subgoal index, indicating the currently active phase of the overall plan.
    \item \(r_t\): the dense step reward, providing immediate feedback on the quality of \(a_t\).
    \item \(r^{o}\): the sparse outcome reward, reflecting the final task-level result $R_{\mathrm{final}}$.
\end{enumerate}

\paragraph{Subgoal-Conditioned GRPO for CK-Pro.}
We optimize the main agent using a subgoal-conditioned variant of Group Relative Policy Optimization (GRPO), in which local subgoal-level feedback and global task-level feedback are normalized separately and then combined into a unified step-wise advantage.

\begin{enumerate}
    \item \textbf{Trajectory sampling.}

    Given the current policy \(\pi_{\theta_{\mathrm{old}}}\), we sample a group of \(G\) trajectories, 
    \(\{ \mathcal{T}^{(1)}, \dots, \mathcal{T}^{(G)} \}\). 
    Each trajectory \(i\) has length \(T_i\) and is represented as 
    \[
    \mathcal{T}^{(i)} = \{ h_0^{(i)}, h_1^{(i)}, \dots, h_{T_i}^{(i)} \}.
    \]

    \item \textbf{Decoupled normalization.}

    To prevent high-variance final rewards from overwhelming informative local signals (i.e., reward collapse), we normalize the two objectives independently. \(R_{\mathrm{sub}}^{(i, k)}\) denotes the cumulative subgoal reward for trajectory \(i\) in phase \(k\), and \(R_{\mathrm{final}}^{(i)}\) denotes the final task reward for trajectory \(i\).

    \textbf{Stream A: Subgoal-level advantage} \((A_{\mathrm{sub}})\). 

    The subgoal reward is compared against the group of trajectories within the same subgoal phase \(k\). Let \(\mathcal{G}_k \subseteq \{1, \dots, G\}\) denote the subset of trajectories that reach subgoal phase \(k\). We compute the phase-wise statistics across trajectories as:
    \[
    \mu_{\mathrm{sub}, k} = \frac{1}{|\mathcal{G}_k|} \sum_{i \in \mathcal{G}_k} R_{\mathrm{sub}}^{(i, k)}, 
    \quad 
    \sigma_{\mathrm{sub}, k} = 
    \sqrt{
        \frac{1}{|\mathcal{G}_k|} 
        \sum_{i \in \mathcal{G}_k} 
        \left( R_{\mathrm{sub}}^{(i, k)} - \mu_{\mathrm{sub}, k} \right)^2
    }.
    \]
    For any step \(t\) assigned to phase \(k_t\), the corresponding normalized subgoal-level advantage is
    \[
    A_{\mathrm{sub}, t}^{(i)} = 
    \frac{
        R_{\mathrm{sub}}^{(i, k_t)} - \mu_{\mathrm{sub}, k_t}
    }{
        \sigma_{\mathrm{sub}, k_t} + \epsilon
    }.
    \]

    \textbf{Stream B: Final task-level advantage} \((A_{\mathrm{final}})\). 

    In contrast, the final task reward is normalized using global group statistics across all \(G\) trajectories:
    \[
    \mu_{\mathrm{final}} = \frac{1}{G} \sum_{i=1}^{G} R_{\mathrm{final}}^{(i)}, 
    \quad 
    \sigma_{\mathrm{final}} = 
    \sqrt{
        \frac{1}{G} 
        \sum_{i=1}^{G} 
        \left( R_{\mathrm{final}}^{(i)} - \mu_{\mathrm{final}} \right)^2
    }.
    \]
    The resulting final task-level advantage is
    \[
    A_{\mathrm{final}}^{(i)} = 
    \frac{
        R_{\mathrm{final}}^{(i)} - \mu_{\mathrm{final}}
    }{
        \sigma_{\mathrm{final}} + \epsilon
    }.
    \]

    \item \textbf{Advantage composition} \((A_t)\).

    We combine the local tactical advantage and the global strategic advantage at step \(t\) as
    \[
    A_t^{(i)} = A_{\mathrm{sub}, t}^{(i)} + \lambda A_{\mathrm{final}}^{(i)}.
    \]
    Here, \(\lambda\) is a hyperparameter that controls the trade-off between subgoal execution and overall task alignment.

    \item \textbf{Policy optimization.}

    We update the main agent policy \(\pi_\theta\) by maximizing the GRPO surrogate objective \(J_{\mathrm{GRPO}}(\theta)\). For each token \(x\) in the action sequence \(a_t^{(i)}\), we define the importance ratio \(\rho_\theta(x) = \frac{\pi_\theta(x \mid s_{t,<x})}{\pi_{\theta_{\mathrm{old}}}(x \mid s_{t,<x})}\). The objective is formulated as:
    
\[
\begin{aligned}
J_{\mathrm{GRPO}}(\theta) = 
\frac{1}{G} \sum_{i=1}^{G} \sum_{t=0}^{T_i} \sum_{x \in a_t^{(i)}} 
\Bigl[ & \min \left( \rho_\theta(x) A_t^{(i)}, \, \operatorname{clip}\left(\rho_\theta(x), 1-\epsilon_c, 1+\epsilon_c\right) A_t^{(i)} \right) \\
& - \beta \mathbb{D}_{\mathrm{KL}}\left(\pi_\theta(\cdot \mid s_{t,<x}) \,\|\, \pi_{\mathrm{ref}}(\cdot \mid s_{t,<x})\right) \Bigr],
\end{aligned}
\]

    where \(\epsilon_c\) is the clipping margin and \(\beta\) is the KL penalty coefficient. The KL penalty can be estimated per-token using an unbiased estimator, preventing the policy from deviating excessively from the reference model.
\end{enumerate}

\subsection{Inference-time Scaling}
\label{sec:reflection}

We introduce two inference-time optimization procedures---\emph{reflection with verification} and \emph{voting}---that allow the agent to evaluate and refine its own trajectories, improving robustness and accuracy.

\paragraph{Reflection with Verification}

Reflection allows the agent to review its previous actions after each task attempt. The reflection module summarizes the full trajectory in an action--observation format (e.g., ``Action 1: \ldots, Observation 1: \ldots''). To obtain structured evaluation signals, we use DeepVerifier~\citep{wan2026verifier}, a rubric-based verifier built on an automatically constructed failure taxonomy. Specifically, DeepVerifier evaluates both the trajectory and the predicted answer along five dimensions: \textbf{finding sources} (use of specific and authoritative evidence), \textbf{reasoning} (logical consistency and faithfulness to evidence), \textbf{problem understanding and decomposition} (correct interpretation of the task and subgoals), \textbf{action execution} (correct use of tools, formats, and modalities), and \textbf{trajectory efficiency} (ability to reach a valid answer within the step budget). Each dimension is rated on a four-level scale: \emph{excellent}, \emph{good}, \emph{needs improvement}, or \emph{poor}.

If the agent identifies any violations of these criteria, it will attempt the task again with additional feedbacks: comprising per-rubric scores and targeted error localization that pinpoint which failure category was triggered and at which trajectory step. The agent repeats this process until a satisfactory answer is produced or a predefined retry limit is reached.

\paragraph{Voting}
The voting process enables the agent to aggregate multiple trajectories, enhancing its decision-making and increasing the likelihood of achieving optimal outcomes. In practice, the agent attempts the same task several times, summarizes all resulting trajectories, and then selects the trajectory that best adheres to the rubric-based guidelines established in the reflection process as the final output. DeepVerifier assigns a structured score to each trajectory based on actively retrieved web evidence, and the candidate with the highest verifier score is retained. Unlike reflection, which evaluates each attempt in isolation, the voting process allows the agent to compare and contrast information across multiple trajectories. This comparative approach helps the agent identify higher-quality solutions by leveraging differences among the attempts. For example, when the agent is asked to find a singer's earliest album, one attempt might return an album from the 2000s while another finds one from the 1990s. By comparing these results, the voting module can recognize that the album from the 1990s is the more accurate answer, as it is earlier.

\subsection{Experiments}

Our RL implementation is developed using verl\footnote{https://github.com/verl-project/verl} as the core RL library, with the agent runtime deployed on GPU clusters. Given the volatility of web environments, we specifically utilize Browserless\footnote{https://www.browserless.io} for remote Playwright execution to maximize the stability of web navigation during training.

Table~\ref{tab:ckpro_rl_main} presents the performance of CK-Pro RL on GAIA-text and xbench-DeepSearch. On GAIA-text, the SFT baseline achieves an average Pass@3 of 47.55\%. Although the vanilla RL checkpoint was affected by the volatility of the web-agent environment during experiments, reinforcement learning demonstrates clear advantages when combined with inference-time scaling. Specifically, Subgoal-GRPO achieves an average Pass@3 of 52.67\%, which further improves to 54.61\% when integrated with the verifier. 
The gains are particularly notable on Level 3 tasks, where Subgoal-GRPO Pass@3 + Verifier reaches 41.67\%, significantly outperforming both the SFT Pass@1 (8.33\%) and Pass@3 (16.70\%) results. On xbench-DeepSearch, the benefits of RL are even more pronounced: Subgoal-GRPO + Verifier improves the Pass@1 from 20.00\% to 45.00\% over the SFT baseline, while RL + Pass@3 + Verifier achieves a final score of 58.00\%. These results demonstrate that RL substantially enhances the agent's ability to solve challenging deep research tasks, with performance gains further amplified by inference-time scaling. The comparison between GRPO and our Subgoal-GRPO also demonstrates the effectiveness of finer-grained credit assignment, grounded by milestone achievement scores.

\begin{table}[t]
\centering
\small
\begin{tabular}{l|cccc|c}
\toprule
\midrule
\multirow{2}{*}{\textbf{Deep Research Agent}} & \multicolumn{4}{c|}{\textbf{GAIA-text}} & \multirow{2}{*}{\textbf{xbench-DeepSearch}} \\
& \textbf{Level 1} & \textbf{Level 2} & \textbf{Level 3} & \textbf{Avg.} & \\
\midrule
\textbf{Cognitive Kernel-Pro SFT} & 56.40 & 42.30 & 8.33 &  43.68 & 20.00 \\
-pass@3 & 61.50 & 44.20 & 16.70 & 47.55 & 32.00 \\
\midrule
\textbf{Cognitive Kernel-Pro RL}\footnote{Performance of the initial RL checkpoint} & 50.85 & 30.45 & 13.89 & 36.24 & 20.00 \\
- GRPO + Verifier & 58.97 & 34.61 & 12.50 & 41.25 & 38.00 \\
- Subgoal-GRPO + Verifier & 60.47 & 33.65 & 12.50 & 41.34 & 45.00 \\
- GRPO pass@3 & 58.97 & 48.07 & 25.00 & 49.51 & 56.00 \\
- Subgoal-GRPO pass@3 & {63.72} & {46.92} & {41.67} & {52.67} & {56.00} \\
- Subgoal-GRPO pass@3 + Verifier & \textbf{68.46} & \textbf{47.21} & \textbf{41.67} & \textbf{54.61} & \textbf{58.00} \\
\midrule
\bottomrule
\end{tabular}
\caption{Performance of CK-Pro RL on GAIA-text ($n$=103) and xbench-DeepSearch ($n$=100), reported using Pass@1 and Pass@3.}
\label{tab:ckpro_rl_main}
\end{table} 
\footnotetext{During our experiments with CK-Pro RL, the web-agent environment deteriorated due to a firewall policy update to Bing Search, causing the baseline performance to decline notably. Nevertheless, CK-Pro RL still substantially outperforms the SFT baseline on GAIA-text Pass@3.}

\section{Related Work}

\subsection{Deep Research Agents}
The field of deep research agents has rapidly evolved, driven by the need for autonomous systems capable of conducting complex, multi-step research tasks. These agents leverage large language models (LLMs) and vision-language models (VLMs) to perform tasks such as web navigation, data analysis, code generation, and report synthesis. Below we introduce both close-source and open-source deep research agents.

\paragraph{Proprietary Deep Research Agents}

Proprietary systems have set a high standard for deep research agents by demonstrating robust performance in autonomous task execution. \textbf{OpenAI's Deep Research}~\citep{deep_research} integrates most advanced OpenAI models to autonomously browse the web, analyze data, and generate comprehensive reports. Powered by a specialized version of the o3 model, it achieves strong performance on benchmarks like GAIA (67.36\% average pass@1 accuracy, 72.57\% cons@64 accuracy) and Humanity’s Last Exam (26.6\% accuracy), significantly outperforming other models.
\textbf{Google's Gemini Deep Research}~\citep{gemini_deep_research} was part of the Gemini 2.5 suite, it autonomously searches hundreds of websites, reasons iteratively, and produces detailed reports, emphasizing real-time adaptability and multimodal processing.
\textbf{Perplexity's Deep Research}~\citep{perplexity_deep_research} excels in domains like finance, marketing, and technology, achieving 21.1\% accuracy on Humanity’s Last Exam and 93.9\% on SimpleQA. 
It iteratively searches, reads documents, and refines research plans. 
More recent work, \textbf{Kimi-Researcher}~\citep{kimi_k2_github, kimi_researcher}, an advanced feature of Moonshot AI's Kimi platform, excels in delivering precise research outputs for complex queries across diverse domains.

\paragraph{Open-Source Deep Research Frameworks}

Open-source frameworks have made significant strides in democratizing deep research agents, with notable contributions including Hugging Face's SmolAgents~\citep{smolagents}, a lightweight Python library that supports various LLMs for web search and data processing but may lack optimization for complex, multi-step research tasks; Alibaba Tongyi's WebAgent Framework, comprising WebDancer~\citep{webdancer}, WebSailor~\citep{websailor}, and WebShaper~\citep{tao2025webshaper}, which excels in super-human reasoning for web-based tasks like BrowseComp, GAIA, and WebWalkerQA~\citep{WebWalker}. 
OWL (Optimized Workforce Learning; \citealp{owl}), a hierarchical multi-agent system that leads open-source frameworks with a 69.09\% average score on the GAIA benchmark, supporting tools for online search, multimodal processing, browser automation, document parsing, and code execution.
TapeAgent~\citep{bahdanau2024tapeagents}, from ServiceNow, uses a ``tape'' log to streamline LLM agent development, matching GPT-4o in tasks like form-filling with cost efficiency. AutoAgent~\citep{tang2025autoagent} enables non-technical users to create LLM agents via natural language, achieving 55.15\% GAIA accuracy and excelling in multi-agent tasks \citep{tang2025autoagent}. 
OAgent~\citep{oagent}, an open-source platform, supports modular agent building for reasoning and automation but may rely on proprietary tools.

In all, open-source frameworks for deep research agents lag behind proprietary systems in performance and accessibility. 
Although some open-source agents demonstrate competitive results, they often depend on proprietary tools, limiting their reproducibility. 
Furthermore, research on open-source agent foundation models remains underexplored, as most efforts rely on prompting external APIs. 
In this work, we address these gaps by developing a fully open-source framework and model, leveraging (to the maximum extent) freely available tools to enhance accessibility and performance.

\subsection{Process Credit Assignment and Multi-Reward Optimization in Agentic RL}

Recent GRPO-style reinforcement learning for LLM agents has progressed along two largely complementary directions. One line of work addresses the credit-assignment problem in long-horizon agent trajectories, where supervision is often available only at the final outcome. The other studies how to optimize effectively once multiple intermediate rewards or objectives are already available. This distinction is useful for situating our method: unlike approaches that must infer process supervision indirectly from sparse outcomes, we assume explicit hierarchical subgoal structure and focus on how to preserve these structured signals during optimization.

\paragraph{Process Credit Assignment from Sparse Outcome Rewards.}
A major line of recent work seeks to transform sparse trajectory-level rewards into denser step-level supervision. TreeRPO~\citep{yang2025treerpotreerelativepolicy} does so through explicit tree sampling: for an intermediate reasoning step, its pseudo-reward is estimated from the expected final rewards of descendant branches, avoiding a separate step reward model while providing denser feedback. ARPO~\citep{dong2025agenticreinforcedpolicyoptimization} makes this idea more practical for agentic settings by allocating additional exploration to high-entropy states, especially after tool interactions, and by introducing advantage attribution estimation for stepwise tool-use decisions. AEPO~\citep{dong2025agenticentropybalancedpolicyoptimization} further argues that entropy-guided branching can itself become unstable; it therefore balances rollout allocation across global and branch sampling, penalizes consecutive high-entropy tool-call steps, and modifies the clipping objective with a stop-gradient design to preserve learning signals on high-uncertainty tokens. Orthogonal to tree-based expansion, GiGPO~\citep{feng2025groupingrouppolicyoptimizationllm} remains critic-free and avoids auxiliary reward models by grouping repeated environment states across trajectories through anchor state grouping, enabling micro-level relative advantage estimation for actions taken from the same state. Finally, CSO~\citep{li2026verifiedcriticalstepoptimization} adopts a selective verification-based strategy: instead of scoring every step, it identifies candidate critical steps in failed trajectories, proposes alternative actions with stronger models, and keeps only those alternatives that the current policy can successfully roll out to a correct outcome, thereby constructing high-quality local preference data. Collectively, these methods show several ways to induce process supervision when explicit subgoal rewards are unavailable.

\paragraph{Multi-Reward Optimization with Explicit Intermediate Signals.}
A complementary line of work assumes multiple reward components are already available and studies how to combine them effectively in GRPO-style optimization. MO-GRPO~\citep{ichihara2025mogrpo} introduces variance-based reweighting to prevent over-optimizing easier objectives at the expense of others. Related Multi-Reward GRPO variants~\citep{yixuan2025multirewardgrpo, zhong2025multirewardgrpostableprosodic} combine weighted rule-based and model-based rewards in applications such as text-to-speech and debiasing. However, these methods largely follow a ``weight-then-sum'' or ``sum-then-normalize'' design, which can blur fine-grained supervision when high-variance outcome rewards dominate auxiliary signals. Most relevant to our work is GDPO~\citep{liu2026gdpo}, which identifies a reward-collapse phenomenon in multi-reward GRPO: distinct reward combinations can map to identical advantages when rewards are summed before normalization. By decoupling normalization across reward components before aggregation, GDPO preserves finer-grained signal differences and improves training stability. Our work is closest to this line: rather than inferring process rewards from sparse outcomes, we exploit explicit hierarchical subgoal rewards and optimize them without collapsing their distinct training signals.


\section{Conclusion}

In this work, we introduce Cognitive Kernel-Pro, a fully open-source generalist agent framework that maximizes the use of free tools, achieving state-of-the-art performance on the GAIA benchmark among open-source, free-tool agents while remaining competitive with frameworks relying on proprietary tools. 
Additionally, we explore the training of an open-source agent foundation model within this framework, developing an 8B-based model that surpasses previous counterparts such as WebDancer and WebSailor. Furthermore, we improve agent performance through reinforcement learning, leveraging multi-reward optimization with subgoal-conditioned GRPO.
Future efforts will concentrate on advancing more capable, multimodal agent foundation models to address increasingly complex tasks.

\bibliography{ref, agent_ref, agent_self_improve, rl_ref}
\bibliographystyle{colm2024_conference}

\appendix

\clearpage

\section{Technical Details of Cognitive Kernel-Pro Framework}\label{appx:agent_implementation}

\paragraph{Code-based Action and Tool-using.}
Both the main agent and the sub-agents employ a similar multi-step workflow for their problem-solving process. We utilize code-based actions: all actions, including sub-agent and tool invocations, are defined as Python functions. The agents generate code to call these functions, which are then executed directly to perform the corresponding actions (using Python's built-in \texttt{exec} function). To capture the outputs from action execution, we use Python’s built-in \texttt{print} function; these outputs are subsequently fed into later steps to provide intermediate results. Here are the instructions with regard to the action output format:
\begin{tcolorbox}[colback=blue!5!white, colframe=blue!75!black, breakable, title = {\textsc{Action Output Format}}]

\ttfamily
\footnotesize

\#\# Output\\
Please generate your response, your reply should strictly follow the format:\\
$\bullet$ Thought: Provide an explanation for your action in one line. Begin with a concise review of the previous steps to provide context. Next, describe any new observations or relevant information obtained since the last step. Finally, clearly explain your reasoning and the rationale behind your current output or decision.\\
$\bullet$ Code: Output your python code blob for the next action to execute. Remember to wrap the code with markdown python code marks and print your output.

\end{tcolorbox}

\paragraph{State-enhanced Problem-solving Workflow.}
In addition to code-based actions, our workflow incorporates explicit planning and state management. Specifically, before each action decision, the agent adopts a planning step, formulating plans based on previous steps and the latest observations. A crucial mechanism in this process is the maintenance of a progress state, which records summaries of important information from previous steps, including intermediate results and lessons learned from earlier attempts. This progress state offers concise historical context and guides subsequent actions. The following instructions detail the structure of the progress state for the main agent:
\begin{tcolorbox}[colback=blue!5!white, colframe=blue!75!black, breakable, title = {\textsc{Progress State}}]

\ttfamily
\footnotesize

\#\# Progress State\\
The progress state is crucial for tracking the task's advancement and includes:\\
$\bullet$ \textbf{completed\_list} (List[str]): A list of completed steps and gathered information essential for achieving the final goal.\\
$\bullet$ \textbf{todo\_list} (List[str]): A list of planned future steps; aim to plan multiple steps ahead when possible.\\
$\bullet$ \textbf{experience} (List[str]): Summaries of past experiences and notes, such as failed attempts or special tips, to inform future actions.\\
$\bullet$ \textbf{information} (List[str]): A list of collected important information from previous steps. These records serve as the memory and are important for tasks such as counting (to avoid redundancy).\\
Here is an example progress state for a task to locate and download a specific paper for analysis:\\
\{\\
$\phantom{1234}$`completed\_list': [`Located and downloaded the paper (as paper.pdf) using the web agent.', `Analyze the paper with the document agent.'],\\
$\phantom{1234}$`todo\_list': [`Perform web search with the key words identified from the paper.'],\\
$\phantom{1234}$`experience': [],\\
$\phantom{1234}$`information': [`The required key words from the paper are AI and NLP.']\\
\}

\end{tcolorbox}

\paragraph{Unified Multi-module Communication.}
A key aspect of our multi-module system design is the specification of communication between the main agent and the sub-agents. To ensure simple and robust communication, we adopt a unified and minimal text-based interface for all sub-agent calling. Each sub-agent is implemented as a callable function following the protocol below:
\begin{itemize}[leftmargin=*]
    \vspace{-2mm}
    \item \textbf{Input}: The sub-agent accepts an input argument of ``task'', which is a plain string describing the sub-task assigned to it. Optionally, the sub-agent may accept additional arguments specific to its functionality (e.g., file paths for the file agent).
    \vspace{-2mm}
    \item \textbf{Output}: The sub-agent returns a dictionary with two fields: ``output'', a string containing the well-formatted answer that strictly adheres to any specified output format; and ``log'', a string providing supplementary notes, such as steps taken, issues encountered, or relevant context.
    \vspace{-2mm}
    \item \textbf{Definition}: To enable the main agent to understand the utilities and use cases of each sub-agent, all sub-agents provide a Python docstring-style definition, which is provided to the main agent. For example, the definition of the web agent is as follows:
\begin{tcolorbox}[colback=blue!5!white, colframe=blue!75!black, breakable, title = {\textsc{Web-agent Definition}}]

\ttfamily
\footnotesize

def web\_agent(task: str) $\rightarrow$ dict:\\
$\phantom{1234}$""" Employs a web browser to navigate and interact with web pages to accomplish a specific task.\\
$\phantom{1234}$Args:\\
$\phantom{12345678}$task (str): A detailed description of the task to perform. This may include: 1) The target website(s) to visit (include valid URLs); 2) Specific output formatting requirements; 3) Instructions to download files (specify desired output path if needed).\\
$\phantom{1234}$Returns:\\
$\phantom{12345678}$dict: A dictionary with the following structure: `output' (str): The well-formatted answer, strictly following any specified output format; `log'(str): Additional notes, such as steps taken, issues encountered, or relevant context.\\
$\phantom{1234}$Notes:\\
$\phantom{12345678}$- If the `task` specifies an output format, ensure the `output' field matches it exactly.\\
$\phantom{12345678}$- The web agent can download files, but cannot process or analyze them. If file analysis is required, save the file to a local path and return control to an external planner or file agent for further processing.\\
$\phantom{1234}$Example:\\
$\phantom{12345678}$$>>>$ answer = web\_agent(task=`What is the current club of Messi? (Format your output directly as club\_name.)')\\
$\phantom{12345678}$$>>>$ print(answer)\\
$\phantom{1234}$"""

\end{tcolorbox}
\end{itemize}
With these unified input/output definitions, our system can flexibly manage interactions and collaboration between the main agent and sub-agents, facilitating extension to a wide range of processing scenarios.

\section{Details of Agent-Based Data Construction}\label{appx:agent_query_construction_details}

We present the key prompt templates for agent-based data synthesis.

\begin{tcolorbox}[colback=blue!5!white, colframe=blue!75!black, breakable, title = {\textsc{Data Synthesizing Requirements}}]

\ttfamily
\footnotesize

$\bullet$ Source-Based Queries: Each query must be based on verifiable sources of truth (e.g., Wikipedia, arXiv, Papers With Code, GitHub, or a specific downloadable file whose location is unambiguous). Clearly specify the sources within the query to avoid ambiguity.\\
$\bullet$ Cross-Source Reasoning: Combine information from multiple sources to formulate a challenging and interesting query. The answer should require synthesis, not simple lookup.\\
$\bullet$ Novelty Requirement: The answer must not exist verbatim on the internet. Construct queries that require combining facts or data in a way that produces a new, non-trivial answer.\\
$\bullet$ Stable \& Unambiguous Answers: The answer should be a number or at most a few words, concise and unambiguous. Avoid queries whose answers may change over time or due to data updates.\\
$\bullet$ Self-Containment: The query must be fully self-contained, requiring no external context or references beyond what is provided in the query itself. All necessary details must be included to ensure only one correct answer.\\
$\bullet$ Clarity \& Precision: Ensure the query is clear and precise, specifying all necessary details to avoid multiple interpretations. Clearly state the expected answer format within the query.\\
$\bullet$ Minimal Procedural Detail: Do not include step-by-step instructions or detailed procedures in the query. Focus on the information need, not the process.\\
$\bullet$ Annotator Feasibility: The query should be answerable in a reasonable amount of time by a human annotator.\\
$\bullet$ Interest \& Utility: The query should be interesting and useful -- answering it should provide value and demonstrate the assistant's ability to synthesize and reason across sources.\\
$\bullet$ Multi-Ability Requirement: Queries are encouraged to require the agent to use multiple abilities, such as Web Browsing, File Handling and Multi-Modal Processing.
\end{tcolorbox}

\begin{tcolorbox}[colback=blue!5!white, colframe=blue!75!black, breakable, title = {\textsc{Seed Topics}}]

\ttfamily
\footnotesize

Notable open-source projects in natural language processing (GitHub, Papers With Code)\\
The evolution of jazz music in the 20th century (Smithsonian Institution, Wikipedia)\\
Key literary works of the 19th century (Project Gutenberg, Wikipedia)\\
Advances in space exploration since 2000 (NASA, Wikipedia)\\
The history and cultural significance of the Olympic Games (Olympic.org, Wikipedia)\\
Overview of major world languages and their distribution (Ethnologue, Wikipedia)
\end{tcolorbox}

\begin{tcolorbox}[colback=blue!5!white, colframe=blue!75!black, breakable, title = {\textsc{Hint-based Query Augmentation}}]

\ttfamily
\footnotesize
\{Original\_Query\}\\
<secret>\\
Below are some confidential hints for your reference:\\
\{Hint\}\\
Important Instructions:\\
$\bullet$ Do not disclose or imply in any way that you have access to these hints during your problem-solving or reasoning process.\\
$\bullet$ A strict evaluator will review your entire solution. If your output suggests you relied on these hints, you will be disqualified from your role as a problem-solving agent.\\
$\bullet$ For any sub-problems where you do not know the answer, continue to use appropriate tools and sub-agents as if you are unaware of the hints.\\
$\bullet$ If there is a conflict between information obtained from your tools and the provided hints, always prioritize the information from your tools.\\
$\bullet$ Do not attempt to plan everything in advance or act as if you have privileged foresight.\\
$\bullet$ Remember, maintaining this role is crucial -- do not risk your position by revealing or depending on the hints.\\
Proceed with utmost caution and professionalism.\\
</secret>
\end{tcolorbox}

\end{document}